\documentclass{article}

\usepackage{arxiv}
\usepackage{graphicx}
\graphicspath{ {images/} }
\usepackage[utf8]{inputenc} % allow utf-8 input
\usepackage[T1]{fontenc}    % use 8-bit T1 fonts
\usepackage{hyperref}       % hyperlinks
\usepackage{url}            % simple URL typesetting
\usepackage{booktabs}       % professional-quality tables
\usepackage{amsfonts}       % blackboard math symbols
\usepackage{nicefrac}       % compact symbols for 1/2, etc.
\usepackage{microtype}      % microtypography
\usepackage{lipsum}
\usepackage{framed,multirow}
\title{EdgeNet: A novel approach for Arabic numeral classification}

\author{S.M.A. Sharif\qquad Ghulam Mujtaba\qquad S. M. Nadim Uddin\\College of IT, Gachon University, South Korea \\ \texttt{\{sharif, mujtaba, nadim\}@gc.gachon.ac.kr}}

\begin{document}
\maketitle

\begin{abstract}
Despite the importance of handwritten numeral classification, a robust and effective method for a widely used language like Arabic is still due. This study focuses to overcome two major limitations of existing works: data diversity and effective learning method. Hence, the existing Arabic numeral datasets have been merged into a single dataset and augmented to introduce data diversity. Moreover, a novel deep model has been proposed to exploit diverse data samples of unified dataset. The proposed deep model utilizes the low-level edge features by propagating them through residual connection. To make a fair comparison with the proposed model, the existing works have been studied under the unified dataset. The comparison experiments illustrate that the unified dataset accelerate the performance of the existing works. Moreover, the proposed model outperforms the existing state-of-the-art Arabic handwritten numeral classification methods and obtain an accuracy of 99.59\% in the validation phase. Apart from that,  different state-of-the-art classification models have studied with the same dataset to reveal their feasibility for the Arabic numeral classification. Code available at http://github.com/sharif-apu/EdgeNet.

\end{abstract}

% keywords can be removed

\section{Introduction}
Handwritten numeral classification (HNC) is one of the most prominent research areas in computer vision for several decades. HNC is considered as a sub-field of optical character recognition (OCR) and aims to convert handwritten images into computer readable texts \cite{hakro2014issues}.  In general, HNC  has several real-life applications such as reading passports, recognize license-plates, sort postal mails, process bank cheques, and address book identification \cite{sharif2017evil}. Considering the importance of HNC, a noticeable amount of research works has been conducted on Arabic script.

Arabic is known to be the fourth most widely used language script with more than 422 million native speakers. It is being used as an official language in more than twenty-six countries \cite{doush2018novel}. Unfortunately, existing Arabic HNC methods are still inefficient compared to the classification methods of other language scripts \cite{alginahi2013survey,abdleazeem2008arabic}. Existing Arabic HNC methods are not optimized due to two main reasons.  First, the recent Arabic HNC works are focused on homogeneous data samples \cite{abdleazeem2008arabic}. Performance evaluation of these methods in diverse data samples is still due. Second, the recent works \cite{el2016cnn,ashiquzzaman2017handwritten} used stacked CNN inspired by LeNet \cite{lecun2015lenet}. However, it is well known that the network architecture with the residual connection can accelerate the classification performance for even complex data samples \cite{he2016deep}. Moreover, performance evaluation of the existing state-of-the-art network architectures like VGG, ResNet, DenseNet etc. in Arabic HNC is still unexplored.

The limitations of the existing work have inspired this study to propose a robust and effective method for Arabic HNC. Here, robust and effective are defined as learning from diverse data samples and utilizing available knowledge with an efficient deep model. Hence, this work proposes to unify the existing Arabic numeral datasets and augment them for data diversity (robust) followed by a novel deep network with the low-level features (edge) as residual connection (effective).  Particularly, edge information has been used in different computer vision applications such as object detection \cite{zitnick2014edge}, face recognition \cite{gao2002face}, stereo matching \cite{song2018edgestereo}, image synthesizing \cite{sharif2019sparse}, image inpainting \cite{nazeri2019edgeconnect}, etc. So far, none of the existing HNC works utilize the edge features as residual connection \cite{he2016deep}. The proposed model with edge connection is refereed as \textbf{EdgeNet} in this study.

This study contributes to the Arabic HNC as follows:
\begin{enumerate}
  \item Unify and extend (with augmentation) existing Arabic numeral datasets to introduce data diversity.
  \item Propose a novel deep network (EdgeNet), which can utilize the low-level edge features and propagate it with residual connections.
  \item Study the performance of the existing work with the proposed unified dataset.  
  \item Explore different state-of-the-art classification networks for Arabic HNC.
\end{enumerate}

This paper has been organized into five sections.  Section II highlights on related works, section III described the proposed method. Section IV demonstrates results and comparisons. Finally, Section V concludes the study.

\section{Related Works}
 According to the classification approaches, the Arabic HNC works can be divided into three basic categories: 1) Handcrafted features with a linear classifier, 2) Deep Learning, and 3) Hybrid Approach.

 \textbf{Handcrafted features with a linear classifier.} Prior to the widespread usage of deep learning methods, handcrafted features with linear classifier was considered as the state-of-the-art method for numeral classification.  At that period,  \cite{abdleazeem2008arabic}  studied the feasibility of different linear classifier for ANC.  In their study, they had found that gradient features learned SVM can perform better than the other linear classifiers. In the later year, \cite{mahmoud2008arabic} used a Gabor-based features extraction and applied to an SVM for ANC. With their method, they had achieved a validation accuracy of 97.94\% on a private dataset containing 21,120 image samples. In another study, \cite{alkhateeb2014dbn} utilized the discrete cosine transform (DCT) coefficients approach on a dynamic Bayesian network (DBN). They used a public dataset \cite{el2007two} with 70,000 image samples and obtained a validation accuracy of 85.26\%. 
 
\textbf{Deep Learning.} In recent years, the convolutional neural network (CNN) has outperformed the handcrafted feature extraction based methods for HNC. Particularly,  LeNet had demonstrated a significant improvement over handcrafted feature based classification methods. Following the trend, \cite{el2016cnn} applied LeNet for Arabic HNC and used the same dataset as \cite{alkhateeb2014dbn} used in their method. They outperformed their previous study with deep learning and achieved an accuracy of 88.00\%. In the following year, \cite{loey2017deep} utilized deep autoencoder on the same dataset and set new benchmark results with 98.50 \% validation accuracy. At the same year, \cite{ashiquzzaman2017handwritten} also applied a stacked CNN on another public dataset containing 3000 data sample. Unfortunately, due to the lack of training samples, they were unable to outperform other deep learning based methods and obtained a classification result of 97.40 \%. However, in follow up study, \cite{ashiquzzaman2019efficient} had introduced data augmentation and significantly increased trainable parameters. Subsequently, the model was able to demonstrate a satisfactory improvement over the previous approaches and obtained a validation accuracy of 99.40\%. To the best concern, this method used data augmentation for the first time in ANC. However, their method is questionable due to a large number of trainable parameters.  It has been suspected to suffer from an overfitting phenomenon due to lack of insufficient training samples.

\textbf{Hybrid Approach.} Apart from the previous two categories, a hybrid approach has been adopted by \cite{alani2017arabic}. Moreover, they had combined restricted Boltzmann machine (RBM) and CNN for Arabic HNC.  They had used the same dataset as \cite{ashiquzzaman2017handwritten} and achieved an accuracy of 98.59\% in the validation phase.

Despite the potential capability of edge feature as a residual connection, the recent methods on Arabic HNC are still focusing on stacked CNN.  Moreover, existing works are focused on a specific type of data samples. Their conservative approaches restrict them to obtain maximum performance. Thus, this study focuses on overcoming the limitation of the existing works with a robust and effective HNC method.  Moreover,  the various state-of-the-art classification networks have also been studied for finding their feasibility for a widely used language script like Arabic.

\section{Proposed Method}
\begin{figure*}[hbt]
\centering
\includegraphics[width=\textwidth,keepaspectratio]{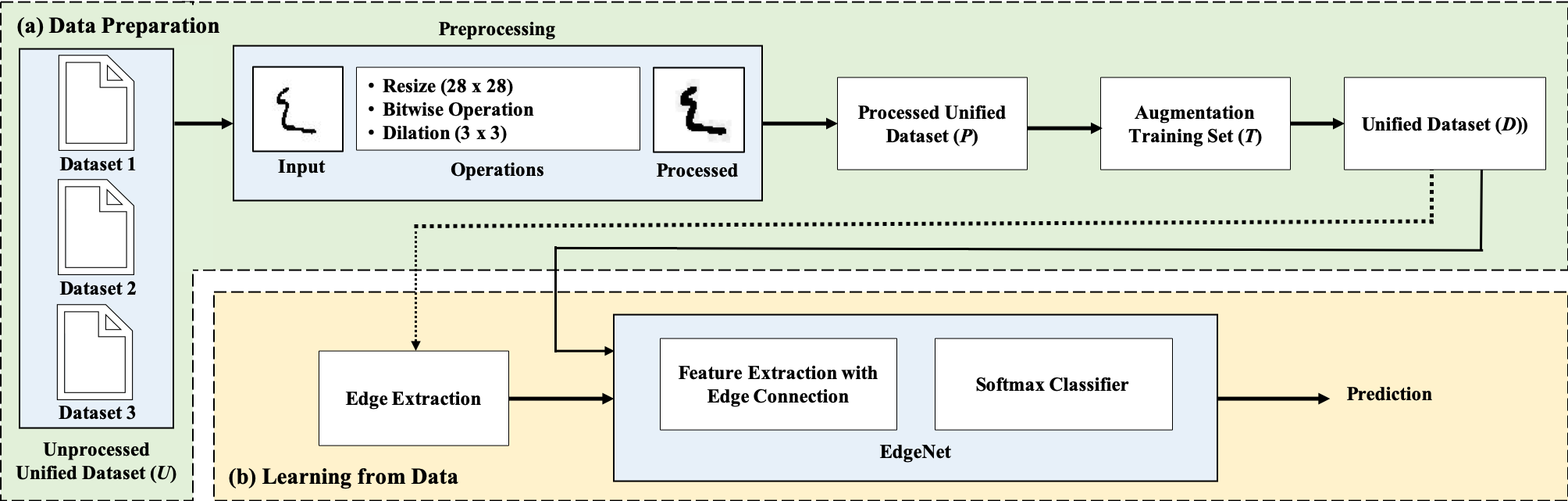}
\caption{ Overview of the proposed work. a) Data Preparation: existing numeral datasets have been merged into a single dataset. The unprocessed unified dataset has been processed to obtain the processed unified dataset ($\mathnormal{P}$). The training set ($\mathnormal{T}$) from the processed unified dataset ($\mathnormal{P}$) has been augmented to introduce data diversity and obtained the unified dataset ($\mathnormal{D}$). b) Learning from data: Image samples from the unified dataset ($\mathnormal{D}$) has been used to extracted edge image. Edge image and the preprocessed images have been used to study the proposed EdgeNet. }
\label{overview}
\end{figure*}

This study proposes a novel method for the Arabic handwritten numeral classification. As Fig. \ref{overview} demonstrates, the proposed method consists of two basic phases: a) data preparation and b)learning from data (EdgeNet). The data preparation method merges the existing Arabic numerals datasets and represents all data into a uniform dataset. Here, the merged dataset with unprocessed data sample has been denoted as the unprocessed unified dataset ($\mathnormal{U}$).  The unprocessed unified dataset ($\mathnormal{U}$) has been processed and presented as the processed unified dataset ($\mathnormal{P}$). The training set from the processed unified dataset ($\mathnormal{P}$) has been enlarged with the augmentation and is denoted as the unified dataset ($\mathnormal{D}$). Later, the image samples from the unified dataset ($\mathnormal{D}$) and their corresponding edge images have been used to study the proposed EdgeNet. Further, the EdgeNet utilizes the inputs and propagate low-level edge feature through a residual connection. The extracted features from the feature extraction block of EdgeNet has been feed into a softmax classifier for the final prediction.

\subsection{Data Preparation}

\begin{figure*}[ht]
\centering
\includegraphics[width=\textwidth,keepaspectratio]{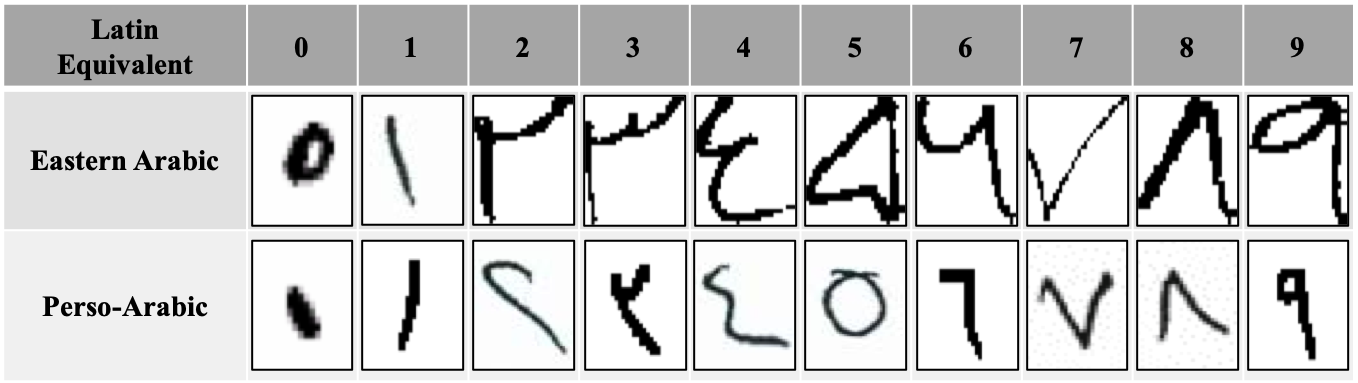}
\caption{ Sample shapes of Arabic handwritten numerals. The top row: Latin equivalent of Arabic numerals, the middle row: handwritten numeral samples of Eastern-Arabic region, the bottom row: handwritten numeral samples of Perso-Arabic region. The used datasets contain more variations of Arabic handwritten numerals.  }
\label{shape}
\end{figure*}

Depending on the region, Arabic script can have different handwritten numeral shapes \cite{khorashadizadeh2016arabic,salimi2013farsi}. Moreover, the handwritten shape of each numeral can be overlapped with another numeral class with different convention e.g. "0" in Eastern Arabic dataset looks almost as same as "5" in Perso-Arabic dataset. As a result, Arabic HNC is considered as a challenging task. Fig. \ref{shape} illustrates samples of the widely used handwritten numeral shapes from the Arabic script.  To utilize the variation of handwritten samples, this study proposes to merge existing Arabic numeral datasets and prepare for the model evaluation. 
\subsubsection{Data Unification}
Three publicly available datasets have been used to prepare the unprocessed unified dataset ($\mathnormal{U}$): PMU-AD \cite{latif2018pmuud}, CMATERDB 3.3.1 \cite{cmaterdb}, and MADBASE \cite{el2007two}. These datasets are collected separately and contain a different number of image samples for an individual image class. In unprocessed unified dataset ($\mathnormal{U}$), the data samples have been shuffled randomly without changing their corresponding image classes. After that, the unprocessed unified dataset ($\mathnormal{U}$) containing 78,180 image samples has been divided into three sub-division as per convention \cite{sharif2019deephog}, where training set ($\mathnormal{T}_U$) contains 62,540 (80\% of overall data), validation set ($\mathnormal{V}_U$) contains 11,712 (15\% of overall data) and testing set ($\mathnormal{C}_U$) set contains 3,928 (5\% of overall data) samples.  Table. \ref{unificationTab} demonstrates the data distribution used in this study. Note that, the processed unified dataset ($\mathnormal{P}$) also incorporated with the same data distribution as the unprocessed unified dataset ($\mathnormal{U}$).

\begin{table*}[ht]
\centering
\begin{tabular}{p{3cm}p{2cm}p{2cm}p{2.1cm}l}
\hline
\textbf{Dataset}         & \textbf{Training Samples ($\mathnormal{T}_U$)} & \textbf{Validation Samples ($\mathnormal{V}_U$)} & \textbf{Testing \newline Samples ($\mathnormal{C}_U$)} & \textbf{Total} \\ \hline
CMATERDB 3.1.1                          &3,000 & - &-                                                          & 3,000           \\ 
PMU-UD                       & 5,180   & - &-                                                         & 5,180           \\ 
MADBase                               & 60,000                     & 10,000 & -                          & 70,000          \\ 
\textbf{Unprocessed}  \newline  \textbf{Unified Dataset ($\mathnormal{U}$)}     & \textbf{62,540 (80\%)}     & \textbf{11,712 (15\%)}       & \textbf{3,928 (5\%)}      & \textbf{78,180} \\ \hline
\end{tabular}
\caption{Overview of used datasets. The proposed unprocessed and processed unified dataset has been obtained from the existing Arabic benchmark datasets (numerals). }
\label{unificationTab}
\end{table*}

\subsubsection{Data Preprocessing}
\begin{figure*}[ht]
\centering
\includegraphics[width=8.5cm,keepaspectratio]{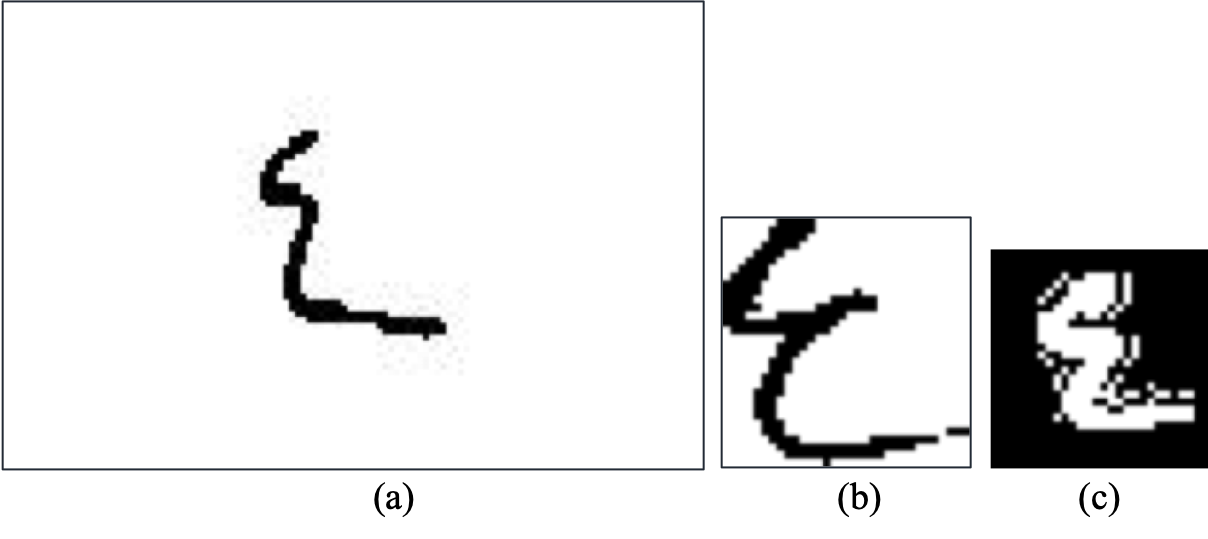}
\caption{ Image samples from the used datasets. a) Sample image from PMU-AD dataset with the dimension of $120 \times 80$. b) Sample image from CMATERDb 3.3.1. dataset with the dimension of $32 \times 32$. c) Smaple image from MADBase dataset with the dimension of $28 \times 28$.  }
\label{sample}
\end{figure*}

 The unprocessed unified dataset $(\mathnormal{U}$) comprises of sample images with various text representations (as they are collected from the different dataset).  Fig. \ref{sample} shows the image samples from used datasets. In PMU-AD dataset, the text appears as black on a white background with an image dimension of $120 \times 80$ pixels. The CMATERDB 3.1.1 dataset has the same text representation as PMU-AD with an image dimension of $32 \times 32$ pixels. The MADBase dataset has a different text representation than previously described datasets. The MADBase dataset adopted the image representation convention suggested by well known MNIST dataset \cite{lecun1998gradient}. Moreover, image samples of MADBase dataset have an image dimension of $28 \times 28$ pixels and the text appears as white on a black background. However, in the proposed unified dataset ($\mathnormal{D}$) all image samples have to be represented in a uniform manner. Hence, the unprocessed unified dataset ($\mathnormal{U}$) has been processed as follows:  $ \mathrm{X} : \mathnormal{U} \to  \mathnormal{P}$. Here,  $ \mathrm{X} $, $\mathnormal{U}$,  and $\mathnormal{P}$ represent a prepossessing function,  unprocessed unified dataset, and processed unified dataset respectively.  As a part of data preprocessing, a sample image $(\mathbf{I}_U$) from unprocessed unified dataset $(\mathnormal{U}$) has been resized into $28 \times 28$ dimension. The background of all images has presented in white through bitwise "not" operation. Although previous study \cite{el2007two} reported that white information on a black background can accelerate the model performance, for the proposed EdgeNet it does not fit well (please see Section 4.1 for details). A morphological dilation \cite{sigmund2007morphology} has been applied with a kernel size of $3 \times 3$ to enhance the information \cite{sharif2019deephog}. Finally, processed unified dataset $(\mathnormal{P}$) has been presented as $\mathnormal{P} = \{\mathnormal{T}, \mathnormal{V},\mathnormal{C} \}$, where $\mathnormal{T}$,$\mathnormal{V}$, and $\mathnormal{C}$ denote the preprocessed training set, preprocessed validation set, and preprocessed testing set. Moreover, each image sample ($\mathbf{I}_P$) of processed unified dataset ($\mathnormal{P}$) has been presented as $\mathbf{I}_P \in [0,1]^{H \times W \times 1}$,  where $H$ and $W$ denote height and width respectively.
 
%\begin{figure}[ht]
%\centering
%\includegraphics[width=9cm,,keepaspectratio]{P.png}
%\caption{ Overview of preprocessing steps. }
%\label{preprocessing}
%\end{figure}

\subsubsection{Data Augmentation}
To the best concern, this method used data augmentation for the first time in ANC. Data augmentation is known to be a useful technique to introduce data diversity in a specific dataset \cite{manjusha2018integrating}. It also helps the deep model to avoid overfitting. In this study, the data augmentation has been intended to introduce data diversity in the available data samples. However, this study does not incorporate into proposing any new data augmentation method. Hence, the augmentation techniques have been adopted as per the suggestion of previous studies \cite{sharif2019deephog,sharif2017evil}. The augmentation has been performed as follows: $ \mathrm{A}: \mathnormal{T} \to  \mathnormal{T}_A$, where $\mathbf{A}$, $\mathnormal{T}$, $\mathnormal{T}_A$  represent the data augmentation method, preprocessed training set and augmented training set respectively.  Here, the augmented training set ($\mathnormal{T}_A$) can be represented as $\mathnormal{T}_A = \{\mathbf{I}_P, \mathbf{I}_R, \mathbf{I}_C, \mathbf{I}_T\}$ , where, $ \mathbf{I}_P, \mathbf{I}_R, \mathbf{I}_C, \mathbf{I}_T$ represent  the preprocessed, rotated, compressed, and translated images respectively.  In later sections, images from $\mathnormal{T}_A$ has denoted as $\mathbf{I}$ for simplicity. The data augmentation has performed as follows:
\begin{itemize}
    \item \textbf{Rotation} : Image ($\mathbf{I}_P$) has been randomly rotated between (-45, +45) to obtain the rotated image ($\mathbf{I}_R$) \cite{sharif2019deephog,sharif2016hybrid}.
    
    \item \textbf{Block Effect}: Image ($\mathbf{I}_P$) has been resized into 14$\times$44 dimension to loss some information\cite{sharif2017evil}.  After that, it has been resized into actual input dimension to obtain a compressed image ($\mathbf{I}_B$). 
    
    \item \textbf{Translation}:  Translated image ($\mathbf{I}_T$) has been obtained by applying a translation matrix ($\mathbf{M}$) as affine translation. In order to find the translation matrix ($\mathbf{M}$), shift direction ($\mathbf{S}_X,\mathbf{S}_Y,$) has been randomly selected between (-5,+5) pixels \cite{sharif2017comparison,sharif2019deephog}. 
\end{itemize}

\begin{figure*}[ht]
\centering
\includegraphics[width=8cm,keepaspectratio]{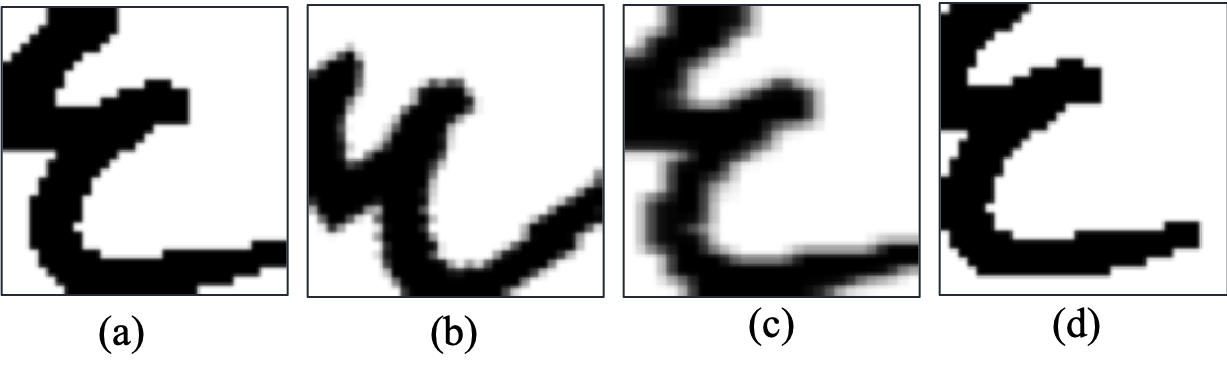}
\caption{ Example of data augmentation. a) Sample image ($\mathbf{I}_P$). b) Rotated image ($\mathbf{I}_R$). c) Blocked image ($\mathbf{I}_B$) d) Translated image ($\mathbf{I}_T$). }
\label{augmentation}
\end{figure*}

Fig. \ref{augmentation} shows a sample image (processed) and its corresponding augmented variants.

\begin{figure*}[ht]
\centering
\includegraphics[width=\textwidth,keepaspectratio]{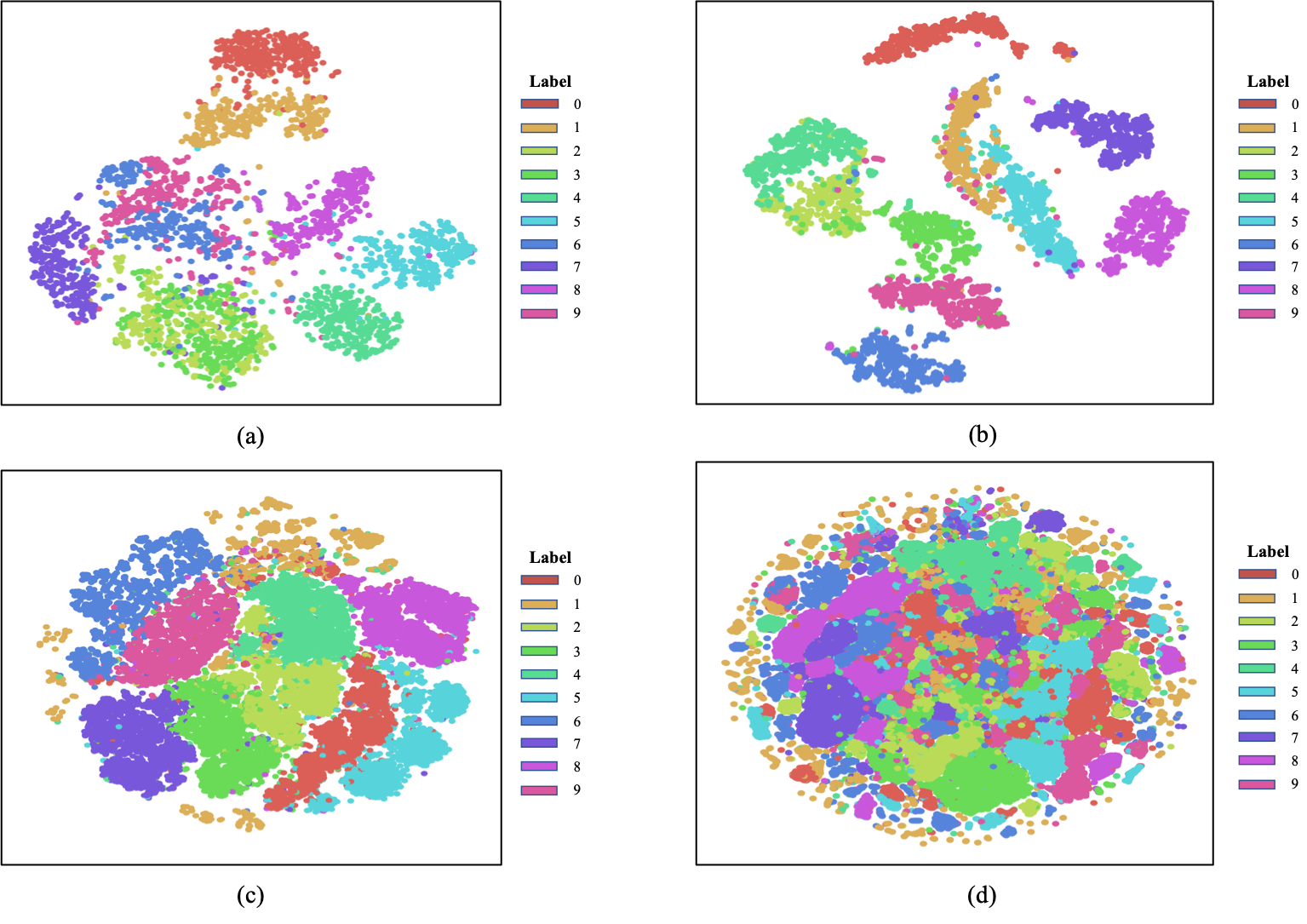}
\caption{  Data distribution of used datasets using t-Distributed Stochastic Neighbor Embedding (t-SNE) (better visual representation can be perceived while viewed in color). The proposed  Unified dataset ($\mathnormal{D}$) is comprised of more scattered distribution comparing to the existing datasets, which aims to help the proposed model to learn more diverse information for real-life applications. a) CMATERDb 3.1.1. b) PMU-UD. c) MADBase. d) Unified dataset ($\mathnormal{D}$).  }
\label{tSNE}
\end{figure*}

The training samples of the training set ($\mathnormal{T}$) has been extended to 250,160 in the augmented training set ($\mathnormal{T}_A$). The processed unified dataset ($\mathnormal{P}$) has been extended to unified dataset ($\mathnormal{D}$) as  $\mathnormal{D} = \{\mathnormal{T}_A, \mathnormal{V}, \mathnormal{C}\}$. The unified dataset ($\mathnormal{D}$) helps this study to introduce data diversity for Arabic HNC. In addition, the diverse data collection aims to prepare the proposed EdgeNet to handle more real-life data samples. As Fig. \ref{tSNE} demonstrates the proposed Unified dataset ($\mathnormal{D}$) contains more scattered data distribution comparing to the existing benchmark datasets. Here, the data distribution has been visualized with t-Distributed Stochastic Neighbor Embedding (t-SNE) \cite{maaten2008visualizing}, which allow intuiting the data arrangement of each dataset in a two-dimensional (2D) space.

\subsection{Learning from Data}
Edge feature and residual connection are two different useful techniques to accelerate performance of a deep model \cite{sharif2019sparse}.  This study proposes to use an edge feature as residual connection for Arabic HNC. Hence, the edge image ($\mathbf{E}$) has been extracted from a sample image ($\mathbf{I}$). Moreover, the proposed EdgeNet has been feed with an input set ($\mathbb{K}$) as $\mathbb{K} = \{\mathbf{I}, \mathbf{E} \}$. It allows the proposed model to learn two different kinds of information simultaneously.
\subsubsection{Edge Extraction}
\begin{figure*}[ht]
\centering
\includegraphics[width=5cm,height=5cm,keepaspectratio]{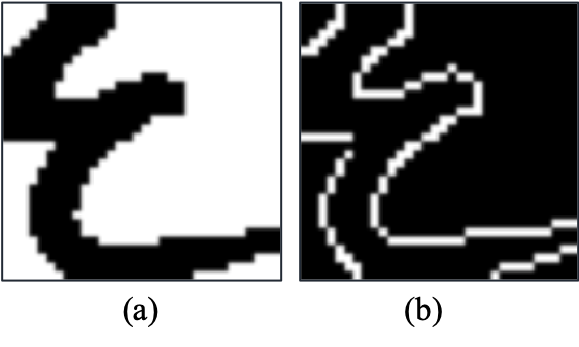}
\caption{  Edge image ($\mathbf{E}$) has been extracted from Sample image ($\mathbf{I}$) through canny edge extraction method. a) Image ($\mathbf{I}$). b) Edge image ($\mathbf{E}$). }
\label{edgeSample}
\end{figure*}

Low-level edge feature has been used in several computer vision applications. Although there are different edge extraction techniques, the canny edge extraction method is known to be useful for the capability of noise suppression and precise shape extraction \cite{zhan2007improved}. Hence, for the proposed EdgeNet, the canny edge extraction method has been used. Edge has been extracted from a sample image ($\mathbf{I}$) as $\mathbf{E} \in [0,1]^{H \times W \times 1}$, where $\mathbf{E}$ is the edge image with height ($H$) and width ($W$). An optimal parameter setting of $minVal=100$, $maxVal=200$, and $apertureSize=3$ has been applied as per the suggestion from a previous study \cite{sharif2019sparse}. Fig. \ref{edgeSample} demonstrates the extracted edge images and their corresponding input samples. The feasibility of the different edge extraction methods has also been demonstrated in a Section. \ref{expEdgeNet}. 

\subsubsection{EdgeNet Architecture}
\begin{figure*}[ht]
\centering
\includegraphics[width=\textwidth,keepaspectratio]{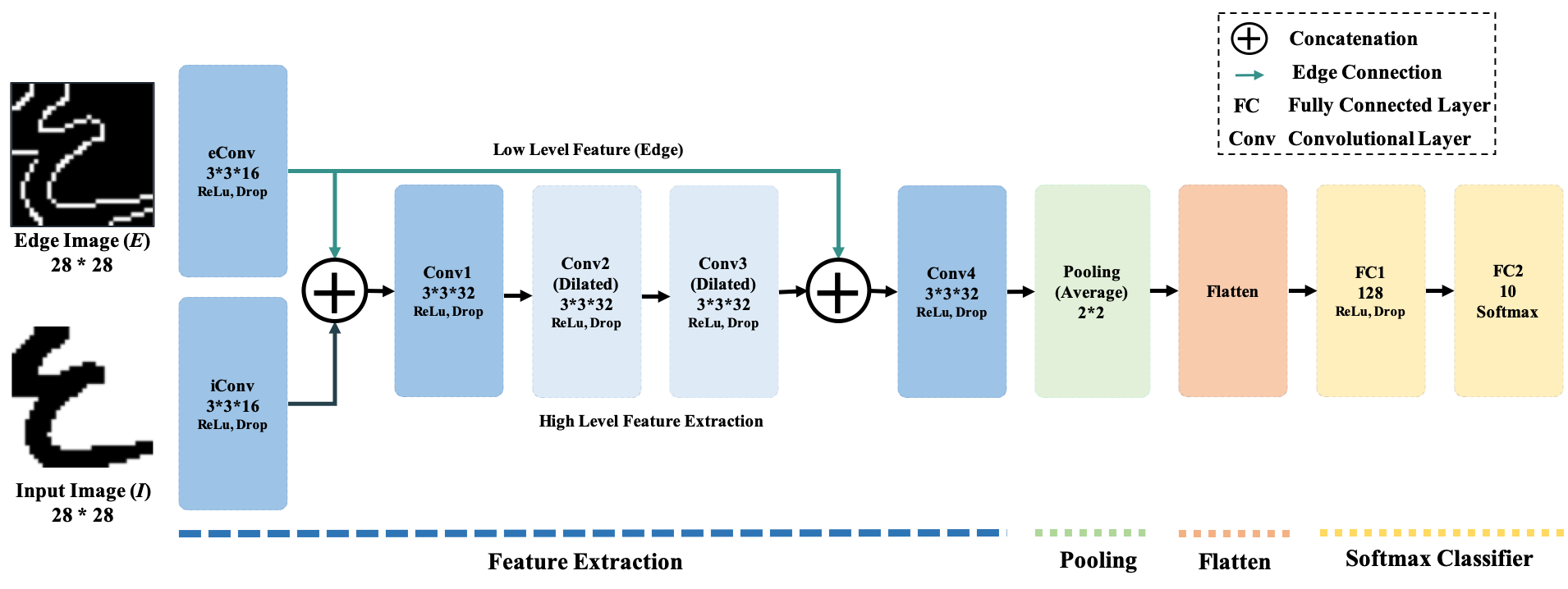}
\caption{ The network architecture of proposed EdgeNet. The low-level edge features have been propagated in the feature extraction block as an edge connection. Extracted features have been fed to a softmax classifier for the final prediction.}
\label{modelArchitecture}
\end{figure*}

As Fig. \ref{modelArchitecture} shows, the proposed model utilizes the input set ($\mathbb{K}$) for the feature extraction with two concurrent convolutional layers (iConv for image feature extraction and eConv for edge feature extraction). Each of these layers comprises a feature depth of 16. The outputs of the iConv and eConv have been concatenated into a single tensor and feed into the Conv1 layer with a depth of 32. After that, two consecutive dilated convolution layers (Conv2 and Conv3) with dilation rate of 2 have been used to extract the high-level features \cite{sharif2019sparse}. The output of Conv3 layer is concatenated with low-level edge feature (eConv) as residual connection and fed into another convolutional layer with feature depth of 32.  This work denotes this residual connection as an edge connection. Here, the edge connection is used to propagate low-level information to the top layer of the feature extraction block. In the feature extraction block, each convolutional layer (including iConv and eConv) is comprised with a kernel size of  $3 \times 3$, strides size of 1, activated with a ReLu \cite{li2017convergence} function and followed by dropout of 25\%. Here, the dropout is used to avoid overfitting \cite{sharif2019deephog}. After the feature extraction, a global average pooling layer has been introduced to optimize the extracted features \cite{boureau2010theoretical}. The pooling layer comprises of a kernel dimension of $2 \times 2$. The output feature vector of the pooling layer has been flattened into a 1D tensor and feed into a softmax classifier. The softmax classifier is comprised of a fully connected (FC1) layer with a dimension of 128 and activated with a ReLu function. Like the feature extraction block, a dropout (rate of 25\%) has been applied after FC1. The FC1 layer has been followed by FC2, which is the final layer of the proposed EdgeNet. The FC2 has the same dimension of image classes and activated with a softmax function \cite{dunne1997pairing}.

\subsection{Experiment Setup}
The trainable parameters of a deep model can be calculated with the following equation \cite{he2015convolutional}:
\begin{equation}
\centering
   \mathnormal{P} = \mathnormal{\sum_{L=1}^{C} {K}_{L-1}.{S}^2_L.{K}_L.{M}^2_L + \sum_{F=1}^{N}(X+1).Y  }  
   \label{paraCompute}
\end{equation}
Here, $\mathnormal{P}$ = total number of parameter, $\mathnormal{C}$ = total number of convolutional layers, $\mathnormal{L}$ = index of convolutional layer, $\mathnormal{{K}_{L-1}}$ = input dimension of convolutional layer, $\mathnormal{S}_L$ =  filter size, $\mathnormal{K}_L$ =  number of filter, $\mathnormal{M}_L$ =  output dimension of convolutional layer, $\mathnormal{F}$ =  index of fully connected (FC) layer, $\mathnormal{N}$ =  number of FC layer, $\mathnormal{X}$ =  input dimension of FC layer, and $\mathnormal{Y}$ =  output dimension of FC layer. Note that the pooling layer has been ignored as it does not add any trainable parameters to a deep model.
As per Eq. \ref{paraCompute}, the total number of trainable parameters in EdgeNet is \textbf{836,938}, which is ${6 times}$ slimmer than the state-of-the-art Arabic HNC work \cite{ashiquzzaman2019efficient} with 4,977,290 trainable parameters. Lower amount of trainable parameters helps the proposed model to train faster and avoid overfitting (please see Section 4 for the details). The proposed model has been trained with an augmented training set ($\mathnormal{T}_A$) and validated with the validation set ($\mathnormal{V}$). The best weight from the validation phase has been preserved and studied with a testing set ($\mathnormal{C}$) for the cross-validation. The learning rate of 0.001 has been selected to control the gradient steps. The batch size has been fixed to 128 and trained for 100 epochs. Adadelta \cite{zeiler2012adadelta} optimizer has been applied to optimize the categorical cross entropy losses. In the later section, the different variants of the proposed model have experimented for the justification with the same hyperparameters (please see section 4.1 for details). All experiment has been conducted on a machine running on Ubuntu 16.04.6 with a hardware configuration of 24GB random-access memory (RAM) and Intel core  i7-7700 central processing unit (CPU). A Titan-XP graphical accelerated processing unit (GPU) has been utilized to accelerate the learning process.

\section{Results and Comparison}
The proposed EdgeNet has been justified with different experiments and compared with existing Arabic HNC works. Finally, the feasibility of the state-of-the-art network architecture from different classification application has been studied for the Arabic HNC.

\subsection{Experiments with EdgeNet Variants}
\label{expEdgeNet}
\begin{figure*}[ht]
\centering
\includegraphics[width=8.5cm,height=5cm,keepaspectratio]{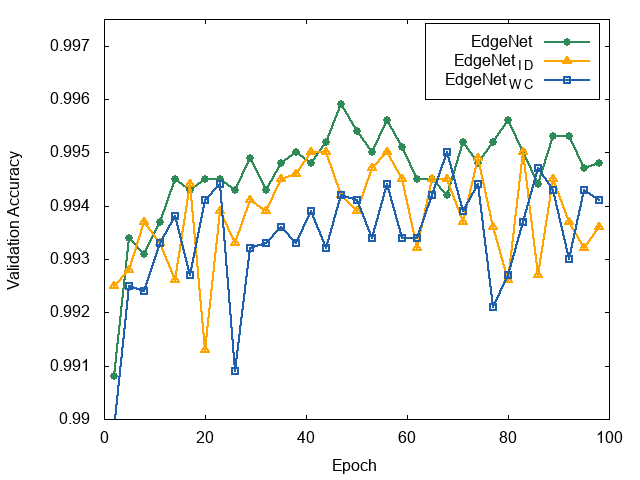}
\caption{ Validation accuracy over training iteration. The proposed EdgeNet demonstrates a consistent validation accuracy over training phase and outperform other variants. }
\label{valAcc}
\end{figure*}

\begin{figure*}[ht]
\centering
\includegraphics[width=8.5cm,height=5cm,keepaspectratio]{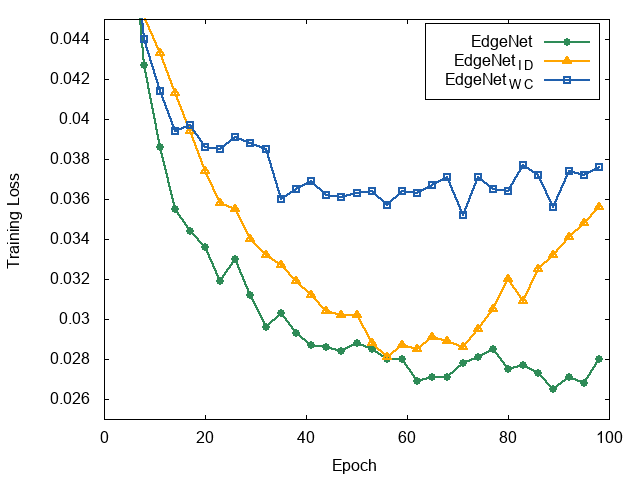}
\caption{ Training loss over training iteration. The proposed EdgeNet with canny edge learn features comparing to other variants.}
\label{trainLoss}
\end{figure*}

The different variants of proposed EdgeNet have been studied by modifying parameters such as data representation, edge extraction method, and removing edge connection. However, the overall architecture of EdgeNet has been remained unchanged. Table. \ref{varient} shows the experiments with EdgeNet variants. Here, EdgeNet$_{WC}$, EdgeNet$_{SE}$, EdgeNet$_{ID}$, and EdgeNet$_{Log}$ denote the variants of EdgeNet, which denote without edge connection,
Sobel edge extraction method,  inverse data, and  Laplacian of Gaussian (LoG) edge extraction method respectively.  The experimental results illustrate that precise shape extraction through canny edge method and propagation of low-level features through edge connection help the proposed EdgeNet to outperform other variants. Moreover, it has obtained the maximum accuracy of 99.59\% in the validation phase. Fig. \ref{valAcc} and Fig. \ref{trainLoss} illustrate the validation accuracy and training loss over training iterations of three EdgeNet variants. Note that only canny edge features have been considered for data visualization as it outperformed other edge features such as Sobel and LoG. From the data visualization, it can be observed that the edge connection with low-level edge features accelerates the learning and validation performance of the proposed EdgeNet. However, the edge images contain white information on black background. Hence, the inverse data (white information on a black background) does not fit well with the proposed EdgeNet. The black background of inverse data samples is inefficient for the proposed EdgeNet as it introduces a tendency of overfitting while training the proposed model.
 
\begin{table*}[hbt]
\centering
\begin{tabular}{l l l}
\hline
\textbf{Network Variant}  & \textbf{Description}                                                                                 & \textbf{Accuracy (Max)} \\ \hline
EdgeNet$_{WC}$        & Edge connection and edge image has been removed.                                                            & 99.51                   \\ 
EdgeNet$_{SE}$                      & Sobel edge $3 \times3$ kernel \cite{gonzalez2016improved}.          & 99.53                    \\ 
EdgeNet$_{ID}$        &  Image background changed into black                                                    & 99.54                   \\ 
EdgeNet$_{LoG}$                  & Laplacian of Gaussian (LoG) with $3 \times3$ kernel \cite{ozturk2015comparison} & 99.57                   \\ 

\textbf{EdgeNet}          & \textbf{Proposed method}                                                         & \textbf{99.59}          \\ \hline

\end{tabular}

\caption{Experiments with EdgeNet variants. The proposed EdgeNet with canny edge extraction method outperforms other variants due to precise edge extraction through the canny edge and low-level edge feature propagation through the edge connection.}
\label{varient}
\end{table*}

\begin{figure*}[ht]
\centering
\includegraphics[width=8.5cm,keepaspectratio]{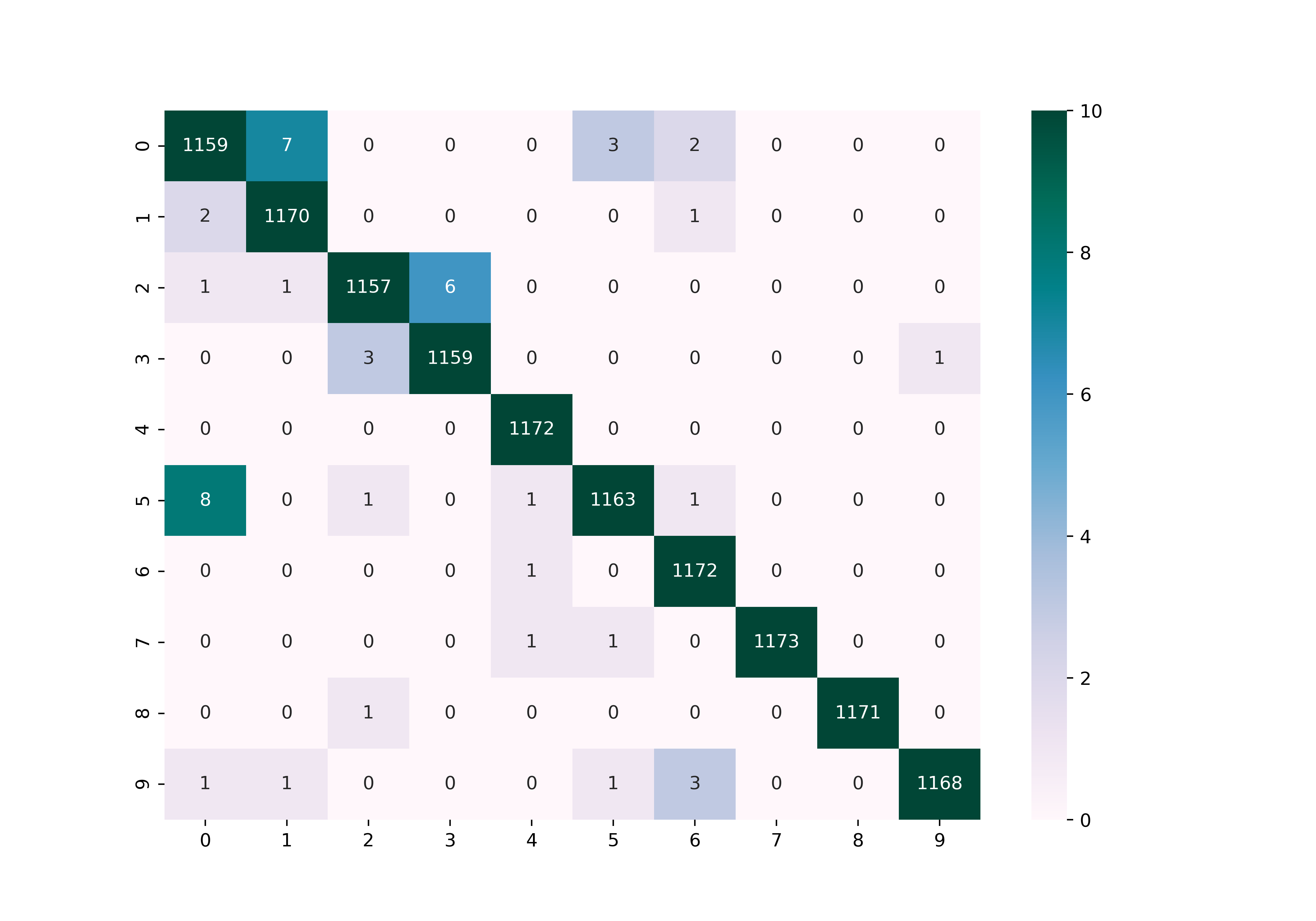}
\caption{ Confusion matrix on the best weight of EdgeNet. Handwritten numerals with similar shape degraded the performance of the proposed EdgeNet. }
\label{confMat}
\end{figure*}

The best weight of proposed EdgeNet has been utilized for further analysis. Moreover, each numeral class has been studied to identify the weak points of the proposed model.  Fig. \ref{confMat} demonstrates that similar shape like $\{0, 1\}, \{0,5\}$, and $\{2, 3\}$ degrade the overall performance of the proposed EdgeNet.

\subsection{Comparison with Existing Arabic Numeral Works}
The existing Arabic HNC works lack of fair comparisons with other works. In most cases, the existing works conduct comparisons with other method incorporated with different datasets and different learning process (e.g., number of iterations). For the fair comparison, every method should be studied with the same dataset and should follow the same learning process. Hence, the existing HNC methods have been studied with the unified dataset ($\mathnormal{D}$). It allows this study to make a fair comparison with the proposed EdgeNet. It also reveals the performance of existing methods in a diverse data collection. As like EdgeNet, the existing methods have been trained with augmented training set ($\mathnormal{T}_A$), validated with the processed validation set ($\mathnormal{V}$), and cross-validated with the processed testing set ($\mathnormal{C}$). The existing models been have tuned with their suggested hyperparameters. Linear classifiers have fed with their suggested handcrafted feature vectors. All models have been trained for 100 iterations (epochs). Table. \ref{sotaWork} shows the method, their reported results, and performance on the unified dataset  ($\mathnormal{D}$). 
\begin{table*}[hbt]
\centering
\begin{tabular}{p{2.2cm}lllll}
\hline

\multirow{2}{*}{\textbf{Method}}              & \multicolumn{2}{l}{\textbf{Reported Result}} 
 & & \multicolumn{2}{l}{\textbf{Result in Unified Dataset}}                          \\  \cline{2-3} \cline{5-6}  
                                              & \multicolumn{1}{c}{\textbf{Reported Accuracy}} & \multicolumn{1}{c}{\textbf{Data Sample}} & & \multicolumn{1}{c}{\textbf{Validation}} & \multicolumn{1}{l}{\textbf{Testing}} \\ \hline
DCN-DBN\cite{alkhateeb2014dbn} & 85.26                                           & 70000                              &       & 92.41                                    & 87.02                                 \\ 
LeNet\cite{el2016cnn}                             & 88.00                                           & 70000    &                                 & 99.12                                    & 99.10                                 \\ 
MLP\cite{ashiquzzaman2017handwritten}                                      & 93.80                                           & 3000                                  &    & 95.57                                    & 88.78                                 \\ 
CNN\cite{ashiquzzaman2017handwritten}                        & 97.40                                           & 3000   &                                   & 99.03                                    & 99.01                                 \\ 
Gabor-SVM\cite{mahmoud2008arabic}                               & 97.94                                           & 21120     &                                & 98.34                                    & 95.99                                 \\ 
RBM-CNN\cite{alani2017arabic}                               & 98.59                                           & 3000    &                                  & 99.02                                    & 98.93                                 \\ 
Autoencoder\cite{loey2017deep}                             & 98.50                                           & 70000        &                             & 98.96                                    & 98.93                                 \\ 
CNN\cite{ashiquzzaman2019efficient}                          & 99.40                                           & 3000   &                                   & 99.40                                    & 99.33                                 \\ 
\textbf{EdgeNet \newline (Proposed)}                              & \multicolumn{2}{c}{\textbf{-}}          &                                                   & \textbf{99.59}                           & \textbf{99.50}                       \\ \hline
\end{tabular}
\caption{Comparison with existing works. The data diversity of unified dataset helps existing methods to accelerate their performance. Moreover, the proposed EdgeNet has utilized data diversity and outperformed the existing works in both validation and testing.}
\label{sotaWork}
\end{table*}

As Table. \ref{sotaWork} demonstrates, the proposed method outperforms the existing works in both validation and testing phase. It achieved a validation and testing accuracy of 99.59\% and 99.50\% respectively. The data diversity of unified dataset ($\mathnormal{D}$) accelerates the performance of the existing methods as well. In both validation and testing phase, the existing methods surpassed their reported results. Particularly, \cite{el2016cnn} method demonstrate a significant improvement (more than 11\%) over their reported result with the unified dataset ($\mathnormal{D}$) due to optimized performance of LeNet. The experiment results also justify that the deep model outperformed the handcrafted feature extraction based methods in a diverse data collection. The deep models are consistent in the testing phase as well. In exception, \cite{ashiquzzaman2019efficient} method does not demonstrate validation improvement on unified dataset ($\mathnormal{D}$) due to a large number of training parameters (\textbf{4,977,290}) and apparent overfitted.

\subsection{Study state-of-the-art Classification Networks on Unified Dataset}
Several state-of-the-art network architectures outperformed the LeNet like simple stacked CNN for image classification. Unfortunately, none of the existing Arabic HNC work explored those network architectures and compared with their respective works. However, this work studied the feasibility of state-of-the-art network architectures for Arabic HNC. All networks have been tuned with their suggested hyperparameters and trained until the model converge with the given data. Table. \ref{sotaNet} illustrates the overview of studied network architectures and their performance on the unified dataset. 
\begin{table*}[hbt]
\centering
\begin{tabular}{ll}
\hline
\textbf{Network Architecture} & \textbf{Accuracy (\%) } \\ \hline
            
ACGAN \cite{odena2017conditional}                 & 98.83                                 \\ 
Hybrid-HOG \cite{sharif2016hybrid}                    & 99.20                                  \\ 

EvilNet \cite{sharif2017evil}                & 99.23                                      \\ 
HRNN \cite{chung2016hierarchical}                      & 99.25                                      \\ 
VGG19 \cite{simonyan2014very}               & 99.27                            \\ 
ResNet \cite{he2016deep}                 & 99.31                             \\ 

VGG16 \cite{simonyan2014very}            & 99.42                             \\ 
InceptionV2 \cite{szegedy2017inception}             & 99.43                               \\
Densenet \cite{huang2017densely}             & 99.47                              \\ 

\textbf{EdgeNet (Proposed)}    & \textbf{99.59}                  \\ \hline
\end{tabular}
\caption{Comparison with state-of-the-art network architectures. The proposed EdgeNet outperforms compared network architectures without introducing a massive number of trainable parameters. }
\label{sotaNet}
\end{table*}

As the Table. \ref{sotaNet} shows, proposed EdgeNet can outperform the existing network architectures with significantly lesser trainable parameters. Apart from that,  It also reveals that the network architectures such as ResNet, DenseNet, VGG, and Inception can outperform the existing Arabic HNC methods.

\subsection{EdgeNet performance on benchmark digit dataset (MNIST)}
The proposed EdgeNet has been optimized particularly for ANC. In general, handwritten numerals of Arabic script is more complex than other language scripts (i.g., English). However, the performance of the proposed EdgeNet has been studied on a benchmark digit dataset known as MNIST. Here, the base MNIST dataset (e.g., without augmentation) has been used to study the performance evaluation. As Table. \ref{sotaNetMNIST} demonstrates, the proposed EdgeNet outperforms other classification models for MNIST dataset. It has obtained an accuracy of 99.55\% in the validation phase.

\begin{table*}[hbt]
\centering
\begin{tabular}{ll}
\hline
\textbf{Network Architecture} & \textbf{Accuracy (\%) } \\ \hline

ACGAN                 & 98.78                                 \\ 

HRNN                       & 98.92                                      \\ 

ResNet                  & 99.21                             \\ 
Hybrid-HOG                  & 99.22                                  \\ 
VGG16             & 99.29                             \\ 
LeNet                   & 99.31                                 \\ 

VGG19                & 99.31   
        \\ 
InceptionV2             & 99.32                               \\ 

Densenet              & 99.39                              \\ 

EvilNet                 & 99.49                                     \\ 
\textbf{EdgeNet (Proposed)}    & \textbf{99.55}                  \\ \hline
\end{tabular}
\caption{Performance evaluation of state-of-the-art network architecture on base MNIST dataset. The proposed EdgeNet demonstrates consistency for MNIST dataset and outperforms existing methods. }
\label{sotaNetMNIST}
\end{table*}

\section{Conclusion}

In this study, a novel approach for Arabic handwritten numeral classification has been proposed. The existing benchmark datasets have been unified and extended with augmentation. It allows this study to introduce data diversity.  Finally,  a deep network with edge connection has been proposed. The experimental results demonstrate that the proposed model with edge connection can outperform the existing state-of-the-art methods for the unified dataset. In the validation phase, the proposed EdgeNet achieved a classification accuracy of 99.59\%. The feasibility of state-of-the-art network architectures is also studied for Arabic HNC. It has been found that different network architectures like VGG, DenseNet, ResNet, Inception etc. can outperform existing Arabic HNC methods. In addition, the proposed EdgeNet can outperform those network architectures without introducing a massive number of trainable parameters. In the foreseeable future, EdgeNet will be studied for more complex data distribution.

\bibliographystyle{unsrt}  
%\bibliography{references}  %%% Remove comment to use the external .bib file (using bibtex).
%%% and comment out the ``thebibliography'' section.

%%% Comment out this section when you \bibliography{references} is enabled.
\bibliography{references}

\end{document}